\documentclass[conference]{IEEEtran}

\ifCLASSINFOpdf
  \usepackage[pdftex]{graphicx}
\else
\fi

\usepackage{amsmath}
\usepackage{url}
\usepackage{xcolor}
\usepackage[bottom]{footmisc}
\definecolor{red}{RGB}{255, 0, 0}

\newcommand\blfootnote[1]{%
  \begingroup
  \renewcommand\thefootnote{}\footnote{#1}%
  \addtocounter{footnote}{-1}%
  \endgroup
}

\begin{document}

\title{AAM-Gym: Artificial Intelligence Testbed for Advanced Air Mobility}

\author{\IEEEauthorblockN{Marc Brittain}
\IEEEauthorblockA{Surveillance Systems\\
MIT Lincoln Laboratory\\
Lexington, MA USA\\
marc.brittain@ll.mit.edu}
\and
\IEEEauthorblockN{Luis E. Alvarez}
\IEEEauthorblockA{Surveillance Systems\\
MIT Lincoln Laboratory\\
Lexington, MA USA\\
luis.alvarez@ll.mit.edu}
\and
\IEEEauthorblockN{Kara Breeden}
\IEEEauthorblockA{Surveillance Systems\\
MIT Lincoln Laboratory\\
Lexington, MA USA\\
kara.breeden@ll.mit.edu}
\and
\IEEEauthorblockN{Ian Jessen}
\IEEEauthorblockA{Surveillance Systems\\
MIT Lincoln Laboratory\\
Lexington, MA USA\\
ian.jessen@ll.mit.edu}}
\maketitle

\begin{abstract}
We introduce AAM-Gym, a research and development testbed for Advanced Air Mobility (AAM). AAM has the potential to revolutionize travel by reducing ground traffic and emissions by leveraging new types of aircraft such as electric vertical take-off and landing (eVTOL) aircraft and new advanced artificial intelligence (AI) algorithms. Validation of AI algorithms require representative AAM scenarios, as well as a fast time simulation testbed to evaluate their performance. Until now, there has been no such testbed available for AAM to enable a common research platform for individuals in government, industry, or academia. MIT Lincoln Laboratory has developed AAM-Gym to address this gap by providing an ecosystem to develop, train, and validate new and established AI algorithms across a wide variety of AAM use-cases. In this paper, we use AAM-Gym to study the performance of two reinforcement learning algorithms on an AAM use-case, separation assurance in AAM corridors. The performance of the two algorithms is demonstrated based on a series of metrics provided by AAM-Gym, showing the testbed’s utility to AAM research.
\end{abstract}

\IEEEpeerreviewmaketitle

\blfootnote{DISTRIBUTION STATEMENT A. Approved for public release. Distribution is unlimited. \textsuperscript{\copyright} 2022 IEEE. Personal use of this material is permitted. Permission from IEEE must be obtained for all other uses, in any current or future media, including reprinting/republishing this material for advertising or promotional purposes, creating new collective works, for resale or redistribution to servers or lists, or reuse of any copyrighted component of this work in other works. This material is based upon work supported by the United States Air Force under Air Force Contract No. FA8702-15-D-0001. Any opinions, findings, conclusions or recommendations expressed in this material are those of the author(s) and do not necessarily reflect the views of the United States Air Force.}
\section{Introduction}
Advanced air mobility (AAM) has the potential to revolutionize transportation through the introduction of new highly automated aircraft to move passengers and cargo within local, regional, inter-regional, and urban environments~\cite{faa_conops}. The Federal Aviation Administration (FAA) and the National Aeronautics and Space Administration (NASA) have both released initial concepts of operations (CONOPS) detailing their vision of AAM in the near-, mid-, and far-term~\cite{faa_conops,nasa_conops}. While near-term operations are expected to mostly leverage existing helicopter infrastructures, mid- to far-term operations require the introduction of new automation in order to safely scale up operations to achieve the envisioned high-density, high-tempo AAM environment.

In a recent study conducted National Academies, eight key challenges to AAM realization were identified; safety, security, social acceptance, resilience, environment impacts, regulation, scalability, and flexibility~\cite{NAP25646}. When considering the introduction of new advanced automation techniques required for AAM such as artificial intelligence (AI), it becomes critical to understand how these new techniques can be addressed in each of these key challenge areas. In other words, a means of addressing the challenges of AI in these key areas is required to adopt emerging AI techniques.

Research for AI in AAM is hindered by the lack of real operational data and scenarios, given the concept of AAM is still being defined. Simulation, however, provides a low-cost way to develop, train, and validate AI algorithms in representative AAM use-cases. In addition, simulation allows for the exploration of edge-case scenarios that may be too dangerous to be performed in the real world.

Many domains outside of air transportation have greatly benefited from standardized simulation testbeds. The Car Learning to Act (CARLA) simulator for autonomous driving is an open source simulator that has become the standard to evaluate and baseline new AI research in autonomous driving~\cite{DBLP:conf/corl/DosovitskiyRCLK17}. In robotics, the OpenAI Gym environment provides a suite of classical control problems to train, evaluate, and compare new AI algorithms to the existing state-of-the-art~\cite{https://doi.org/10.48550/arxiv.1606.01540}. The benefits of these standardized testbeds are twofold: (1) researchers can rapidly develop and prototype new AI algorithms on existing use-cases and (2) researchers can evaluate existing state-of-the-art AI algorithms on new use-cases. In traditional air transportation, high fidelity modeling and simulation has enabled the development and validation of real-time detect and avoid (DAA) systems such as the Airborne Collisions Avoidance System for small uncrewed aircraft systems (ACAS sXu)~\cite{9081631,https://doi.org/10.48550/arxiv.2204.14250}. Given the increase in sUAS in the national airspace, ACAS sXu will be key DAA system for near-term AAM operations. Therefore, the lessons learned from ACAS sXu's modeling and simulation development can pave the way for a standardized AAM simulation testbed.

There are several examples of simulators being developed for AAM, including OneSky~\cite{OneSky}, SimUAM~\cite{yedavalli2021simuam}, UAMToolkit~\cite{doi:10.2514/6.2021-2381}, BlueSky~\cite{BlueSky}, and many others being developed by companies such as Joby and Airbus. While each of the simulators provide a different unique modeling capability, they all lack a standardized architecture to facilitate rapid AI research and development.

In this paper, we propose the AI Testbed for Advanced Air Mobility, hereafter referred to as AAM-Gym. AAM-Gym is built on the philosophy of the OpenAI Gym environment, where use-cases and algorithms can be expanded upon over time, resulting in a standardized ecosystem for researchers in academia, industry, and government to perform AI research in AAM. AAM-Gym is independent of the backend simulator and is demonstrated with the integration of the BlueSky simulator\footnote{\url{https://github.com/TUDelft-CNS-ATM/bluesky}} and UAMToolkit. AAM-Gym is used to study the performance of two reinforcement learning algorithms on an AAM use-case, separation assurance in AAM corridors, to measure the performance of the two algorithms based on a series of metrics provided by AAM-Gym, showing the testbed’s utility to AAM research.

The structure of this paper is as follows. Section~\ref{background} provides a brief overview of artificial intelligence and how artificial intelligence has been applied in air transportation. Section~\ref{aamgym} introduces the
AAM-Gym testbed and provides details on the underlying architecture. The numerical experiments are presented in Section~\ref{experiments}. Section~\ref{results} provides
an analysis of the results. Conclusions are presented in Section~\ref{conclusion}.

\section{Background}
\label{background}
Artificial intelligence is a branch of computer science that encompasses tasks commonly associated with humans, such as sensing, reasoning, and learning. In general, artificial intelligence algorithms often fall into three categories of machine learning: supervised learning, unsupervised learning, and reinforcement learning, each with a unique protocol for training and evaluation.

Supervised learning techniques often encompass tasks such as image classification, predictive modeling, natural language processing, etc. To perform supervised learning, AI algorithms require a well defined dataset consisting of features, as well as the associated target label to learn a mapping between the features and target label. In recent years, supervised learning has seen many breakthroughs in various domains, such as protein folding in biology~\cite{jumper2021highly} and weather nowcasting in meteorology~\cite{ravuri2021skilful}. Supervised learning has also been successfully applied to air transportation problems such as aircraft trajectory prediction~\cite{https://doi.org/10.48550/arxiv.1812.11670} and uncrewed aircraft systems (UAS) trajectory modeling~\cite{doi:10.2514/6.2017-3072}. These examples prove supervised learning can be applied to air transportation problems, however, it is challenging for others to build upon the work without a standardized training and evaluation testbed. 

The second category of machine learning, unsupervised learning, encompasses tasks such as clustering, image segmentation, etc. In unsupervised learning, AI algorithms require a defined dataset of features, but unlike supervised learning, they do not require the target label for training and instead only require the target label for algorithm evaluation. Unsupervised learning has become a popular approach for aircraft trajectory clustering in air transportation to understand aircraft movements and trajectory phases~\cite{olive:hal-02350789, OpenSky19:Python_Toolbox_for_Processing, aerospace8090266}. While the results of unsupervised learning are promising, there still lacks a standardized way to validate unsupervised learning approaches in air transportation, given the lack of validation target labels and community-based data interfaces.

Reinforcement learning (RL) is a more recent field of artificial intelligence where intelligent agents learn optimal behaviors or strategies in an unknown environment to maximize a cumulative reward signal. While supervised and unsupervised learning algorithms are trained using a static dataset, reinforcement learning algorithms are often trained dynamically, where agents interact directly with the environment to obtain training data. Therefore, simulation environments are essential for reinforcement learning algorithms to obtain large amounts of training data and for validating the algorithms in a safe environment. Today, challenging games such as Go, Atari, Warcraft, and, most recently, Starcraft II have been played by RL agents with beyond human-level performance~\cite{silver2016mastering, mnih2015human, vinyals2019grandmaster}. These notable advancements in the AI field demonstrate the capability of computational learning algorithms to potentially augment and facilitate human tasks. Reinforcement learning in air transportation has been applied to many problems such as conflict detection and resolution~\cite{doi:10.2514/6.2021-1952,PHAM2022103463,doi:10.2514/1.I010807}, autonomous separation assurance~\cite{doi:10.2514/1.I010973, 9564466, 9717997}, and collision avoidance~\cite{li2019optimizing}. The challenge, however, is interpreting comparisons of AI algorithm performance when the algorithms are developed in various simulators with different underlying assumptions. Therefore, introducing a standardized simulation testbed is essential to foster innovative AI research in advanced air mobility.

\section{AAM-Gym}
\label{aamgym}

\begin{figure*}[!t]
\centering
\includegraphics[width=7.0in]{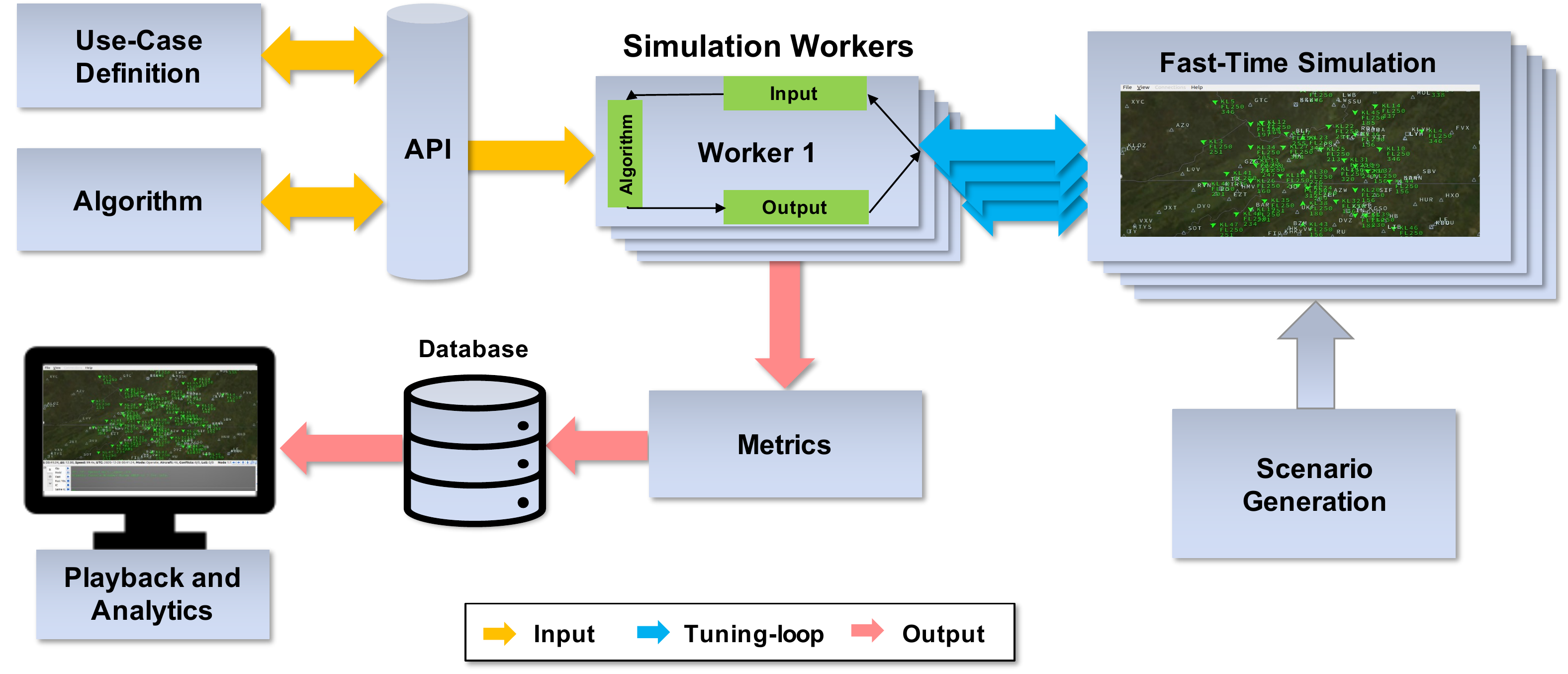}
\caption{Illustration of the AAM-Gym architecture.}
\label{aamgym_architecture}
\end{figure*}

To address the aforementioned gap, MIT Lincoln Laboratory has developed AAM-Gym, the first simulation testbed to facilitate AI research and development in AAM. AAM-Gym was developed from the ground up to streamline the development, training, and validation of AI algorithms on an expandable set of AAM use-cases. The philosophy behind AAM-Gym, based on the community-standard OpenAI gym interface for reinforcement learning~\cite{https://doi.org/10.48550/arxiv.1606.01540}, is to build an extensible set of use-cases and algorithms around a standardized interface. Through a standardized interface, the use-cases define the input-output requirements for the AI algorithms, enabling user's to rapidly test any AI algorithm on any use-case, so long as the AI algorithm follows the OpenAI gym protocol. The architecture of AAM-Gym is illustrated in Figure~\ref{aamgym_architecture}. In this section, the core components of the AAM-Gym architecture are described.

\subsection{Use-Cases}
In order to run a training (or evaluation) experiment in AAM-Gym, users are required to specify: (1) the use-case(s) and (2) the algorithm(s). AAM-Gym use-cases represent fundamental research problems where there are defined objectives, observation spaces, and action spaces. The requirements for the observation space and action space define the input and output dimensions of the AI algorithms, such that the algorithms can be constructed for any use-case. When using AAM-Gym, the user can specify the scale of operations an algorithm needs to address. For example, an AAM macro-level use-case may be strategic flight planning where an AI algorithm determines aircraft flight plans. The objective of this use-case could be to maximize aircraft throughput, minimize delays, etc. It is also possible to operate micro-level use-cases, such as tactical en route collision avoidance where the primary objective is to minimize the number of near mid-air collisions. The difference in these use-cases provides a diverse set of challenges for AI researchers to explore. The flexibility of AAM-Gym provides the means to construct a hierarchical use-case that combines macro-level and micro-level use-cases. For example, joint training and validation of AI algorithms for strategic flight planning and collision avoidance. The combination of these algorithms provides the ability to explore the joint performance of the AI algorithms on overall airspace safety and efficiency. This provides a powerful tool for not only researchers, but also policy makers to help drive important decisions on AAM, for example, structural investments, airspace standards (e.g., corridor dimensions and number of lanes in corridors), and AAM service interoperability requirements. 

\subsection{Algorithms}
AAM-Gym enables users to develop, train, and validate AI algorithms, independent of whether or not the algorithm was developed specifically for AAM. AAM-Gym can ingest algorithms developed for general purpose AI in an effort to rapidly test novel algorithms against AAM use-cases. For example, the Ray RL-Lib~\cite{liang2018rllib} library contains many different state-of-the-art reinforcement learning algorithms optimized for run-time efficiency with the distributed computing backend, Ray~\cite{https://doi.org/10.48550/arxiv.1712.05889}. While these algorithms were not designed for AAM-Gym, they can be used off-the-shelf in AAM-Gym use-cases, providing a quick-look into algorithm performance. This is essential given the pace of artificial intelligence research. While users are free to develop their own AI algorithms for use with AAM-Gym, having the ability to use off-the-shelf algorithms can greatly reduce the time to prototype new algorithms in AAM use-cases.

The algorithm API is based on the OpenAI gym protocol~\cite{https://doi.org/10.48550/arxiv.1606.01540} for single agent use-cases and the RL-Lib protocol~\cite{liang2018rllib} for multi-agent use-cases. While these API protocols are often applied to reinforcement learning algorithms, they can be readily extended to support supervised learning and unsupervised learning since these protocols are simply defining the input-output requirements of the use-cases. AAM-Gym provides examples of algorithm integrations in the popular deep learning modules, PyTorch~\cite{NEURIPS2019_bdbca288} and Tensorflow~\cite{199317}.

\subsection{Configuration}
AAM-Gym was developed with the idea that use-cases and algorithms can be added over time to expand the testbed's research and validation capabilities. With both use-cases and algorithms potentially requiring an exhaustive list of configuration parameters, special care was taken when defining how a user could configure use-cases and algorithms. AAM-Gym leverages the open source Hydra configuration management module~\cite{Yadan2019Hydra} to address hyperparameter traceability, as well as use-case and algorithm configuration. 

Hydra is simple-to-use, well documented, and comes with many convenient off-the-shelf features. These features include: (1) integration with high performance computing clusters, (2) variable type checking, (3) run-time directory management, and (4) automatic hyperparameter value sweeps with job management. Hydra configuration files take the form of YAML files where parameters are listed with their associated value(s). AAM-Gym defines a hierarchical configuration where at the top-level, high-level experiment parameters are defined (e.g., algorithm selection, use-case selection, number of training iterations). At lower levels, the user is free to modify default use-case and algorithm parameters. The convenient feature with Hydra is that all parameters can be modified from the top-level and there is no burden on the user to modify low-level configuration files, further reducing configuration complexity.

\subsection{Distributed Computing}
Given that AI algorithms have high computational workloads, AAM-Gym uses a state-of-the-art distributed computing architecture to enable AI algorithm training and validation on laptops or large supercomputing clusters without any code modifications. This is possible by leveraging the Ray python module for distributed computing~\cite{https://doi.org/10.48550/arxiv.1712.05889}. Ray simplifies distributed computing by managing the inter-process communication and distributed memory, resulting in a simplified API for the developer to conform to. Ray was chosen as the module of choice for distributed computing for two reasons: (1) code written with Ray scales to user's computational resources (e.g., 10's to 1000's of CPUs) and (2) Ray adds minimal complexity to code development allowing developers to focus on new use-cases and algorithms, without spending time on distributed computing. Furthermore, Ray is well documented and maintained, with strong community support.
 
 \subsection{Fast-Time Simulation}
A key requirement for research and validation of AI algorithms is having a suitable fast-time simulation backend. There are many different simulation frameworks that are being explored for AAM, with various fidelity's and capabilities. Rather than identifying a single AAM simulator to use for AAM-Gym, it was determined early on to take a modular approach to backend simulation frameworks. This allows for a wider diversity of AAM-Gym use-cases since not all use-cases may have the same simulation requirements. For example, an emergency landing use-case may require a higher fidelity visualization environment for computer vision algorithms, as well as representative on-board sensors (e.g., LIDAR). Therefore, over time additional backend simulators can be integrated with AAM-Gym to expand the types of use-cases that can be developed.

Currently, AAM-Gym has integrated BlueSky~\cite{doi:10.2514/1.I010973}, which is an open source fast-time air traffic simulator. This simulator provides a suitable fidelity for large-scale airspace use-cases, as well as tactical conflict management use-cases. In future work, additional backend simulation frameworks such as Microsoft AirSim~\cite{https://doi.org/10.48550/arxiv.1705.05065} will be integrated into AAM-Gym.

\subsection{Scenario Generation}
MIT Lincoln Laboratory has developed a suite of tools to enhance the realism of AAM simulations in AAM-Gym. BlueSky does not natively support the generation of scenarios that may be relevant to AAM. Therefore, we have initially focused on three key ares of scenario generation tools to overlay on the backend simulation: (1) integration with the MIT Lincoln Laboratory UAMToolkit~\cite{doi:10.2514/6.2021-2381}, (2) real-world, live air traffic data integration, and (3) real-world representative baselines. By doing this, we are not constrained to a specific backend simulator (i.e., BlueSky), but can use these tools to enhance the realism of other backend simulators as AAM-Gym further develops.

\subsubsection{UAMToolkit Integration}
The MIT Lincoln Laboratory UAMToolKit~\cite{doi:10.2514/6.2021-2381} is composed of a set of demand, scheduling, network design, and queuing models that allow for the evaluation of large scale resource constrained AAM operations. UAMToolkit outputs a complete history of all aircraft operations, such as departure and arrival times, assigned route, cruise speed, altitude, passengers and cargo on-board, origin and destination vertiport, and aircraft type. This output is parsed by AAMGym and translated into a standard BlueSky scenario file to operate the takeoff/landing and cruising phase of each flight.

\subsubsection{Live Air Traffic}
Understanding the interaction of simulated AAM traffic with today's air traffic operations requires the integration of real-world or live air traffic data. While obtaining this data can often be challenging due to restrictions on data access, the OpenSky Network provides free access to ADS-B and Mode-S air traffic data~\cite{6846743}. OpenSky has been shown to be a good data source for research in air transportation due to the ability to collect and process large amounts of data through OpenSky's databases~\cite{9622862}. OpenSky also provides users with python-based API's simplifying the integration with AAM-Gym. This allows real-world air traffic from OpenSky to be overlaid with the simulated AAM traffic to enable more complex research studies.

\subsubsection{Real-World Baselines}
To enable confidence in AI algorithms for safety critical applications such as air transportation, it is critical to ensure proper baselines and scenarios are provided to validate the AI algorithms. Currently, no such testbed has the capability to compare the performance of current state-of-the-art air transportation systems with AI algorithms. However, AAM-Gym provides users the ability to baseline new AI algorithms against their real world system counterpart. For example, a collision avoidance use-case can be baselined against an airborne collision avoidance system (e.g., ACAS sXu) to compare performance. In addition, this provides users a powerful tool to research AI algorithms that augment the performance of existing systems. 

Representative scenarios can be developed manually or automatically generated from data sources. Incorporating AAM traffic demand data and vertiport locations, provides users with realistic scenarios to train and validate AI algorithms. In AAM-Gym, scenarios, and the tools to develop scenarios, are an evolving capability. As new data is provided over time, additional scenarios can be incorporated to meet user’s needs.

\subsection{Metrics}
In order to rapidly validate AI algorithms in AAM-Gym, a suite of metrics are available to the user. For example, in the use-case configuration, a user can specify to collect \textit{safety} metrics. This then grabs all metrics that are associated with the \textit{safety} field in AAM-Gym, such as near mid-air collision (NMAC), loss of well clear (LoWC), etc. If a user specifies \textit{operational}, this will collect metrics related to the operational suitability of the algorithm, such as algorithm alert rate, airborne holding rate, etc. Multiple metrics can be specified as a list (e.g., [\textit{safety}, \textit{operational}]) to allow the user to access all relevant information needed for algorithm validation. Users are also able to add additional metrics as needed, but having a standardized set of metrics provided by AAM-Gym ensures that users report consistent validation metrics.

\section{Numerical Experiments}
\label{experiments}

In this section, the setup of the numerical experiments is described, including the use-case and algorithm definition, along with baselines for algorithm validation.

\subsection{Use-case: separation assurance in AAM corridors}
\subsubsection{State Space}
Use-cases define the input and output requirements for algorithms in AAM-Gym. For reinforcement learning algorithms, this equates to defining the state space, action space, and reward function. It is critical to ensure that the developed use-case aligns with the objectives that the algorithm is trained to achieve. In this use-case, the task of separation assurance is distributed to each aircraft in a decentralized setting, as introduced in~\cite{doi:10.2514/1.I010973}. Therefore, each aircraft from its own viewpoint will have a state associated with itself (ownship state) and the surrounding air traffic (intruder state). The ownship and intruder state information at time $t$ is defined as
\begin{multline*}
    s^{o}_{t} = \\
    (hdg^{(o)}, z^{(o)}, a^{(o)}, v^{(o)}, d^{(o)}_{\text{dest}}, x^{(o)}_{\text{wpt}}(j)-x^{(o)}, y^{(o)}_{\text{wpt}}(j)-y^{(o)}) \\ \forall \; j \in [1, N_{\text{wpt}}],
\end{multline*}
\begin{multline*}
    h^{o}_{t}(i) = \\
    (x^{(i)}-x^{(o)}, y^{(i)}-y^{(o)}, z^{(i)}- z^{(o)}, a^{(i)}, v^{(i)}, d^{(i)}_{\text{dest}},d^{(i)}_{o}, \\
    x^{(i)}_{\text{wpt}}(j)-x^{(o)}, y^{(i)}_{\text{wpt}}(j)-y^{(o)}) \\ \forall \; j \in [1, N_{\text{wpt}}],
\end{multline*}
where $s^{o}_{t}$ represents the ownship state information and $h^{o}_{t}(i)$ represents the state information for intruder $i$ that is available to the ownship at time $t$. The ownship state information includes heading ($hdg^{(o)}$), altitude ($z^{(o)}$), horizontal acceleration ($a^{(o)}$), speed ($v^{(o)}$), distance to the destination vertiport ($d^{(o)}_{\text{dest}}$), and $N_{\text{wpt}}$ future ownship relative waypoint positions ($x^{(o)}_{\text{wpt}}(j)-x^{(o)}$, $y^{(o)}_{\text{wpt}}(j)-y^{(o)}$). The state information for each intruder aircraft includes the ownship relative easting and northing position ($x^{(i)}-x^{(o)}$, $y^{(i)}-y^{(o)}$), relative altitude ($z^{(i)}-z^{(o)}$), horizontal acceleration ($a^{(i)}$), speed ($v^{(i)}$), distance to the destination vertiport ($d^{(i)}_{\text{dest}}$), the straightline distance between ownship and intruder ($d^{(i)}_{o}$), and $N_{\text{wpt}}$ future ownship relative waypoint positions ($x^{(i)}_{\text{wpt}}(j)-x^{(o)}$, $y^{(i)}_{\text{wpt}}(j)-y^{(o)}$). $N_{\text{wpt}}$ is a hyperparameter that is specified in the use-case configuration. This state space definition improves upon the definition in \cite{doi:10.2514/1.I010973} by removing the location dependency, resulting in a state space independent of geographically referenced position.

Not all AI algorithms can support a variable number of state inputs, so a max number of intruder aircraft ($N_{\text{intruders}}$) is provided as a use-case hyperparameter, such that the total state provided to a given ownship is represented as:
\begin{equation*}
    \bar{s}^{o}_{t} = [s^{o}_{t}, h^{o}_{t}(1), ..., h^{o}_{t}(N_{\text{intruders}})].
\end{equation*}
This ensures that the algorithm state size is fixed. In the event that there are fewer aircraft in the simulation than $N_{\text{intruders}}$, the total state is padded with zeros in order to maintain the fixed-length state size.

\subsubsection{Action Space}
The action space follows the definition in \cite{doi:10.2514/1.I010973}, by representing actions as in-trail acceleration advisories bounded by the aircraft performance envelope. The action space is defined as
\begin{equation*}
a_{t} = [a_{-},0,a_{+}],
\end{equation*}
where $a_{-}$ represents deceleration (decrease speed), $0$ represents no acceleration (hold current speed), and $a_{+}$ represents acceleration (increase speed).

\subsubsection{Reward Function}
The reward function reflects the objective of the use-case and provides a signal for the algorithms to optimize. In this use-case, the primary objective is separation assurance, however, secondary objectives are introduced to encourage operational suitability in an AAM environment. The reward function is defined as
\begin{equation}
\label{reward_func}
R(s^{o}_t,h^{o}_t,a_{t}) = R(s^{o}_{t}, h^{o}_{t}) + R(a_{t}) - \Omega,
\end{equation}
where $R(s^{o}_{t}, h^{o}_{t})$ and $R(a_{t})$ are defined as
\begin{equation}
R(s^{o}_{t}, h^{o}_{t}) = 
    \begin{cases}
      -1 & \text{if $d^{c}_{o} < d^{\text{NMAC}}$}\\
      -\alpha + \delta \cdot d^{c}_{o} & \text{if $d^{\text{NMAC}} \leq d^{c}_{o} < d^{\text{MAX}_{i}}_{o}$}\\
      0 & \text{otherwise}
    \end{cases},  
\end{equation}
\begin{equation}
R(a_{t}) = 
    \begin{cases}
      0 & \text{if $a_{t} =$ `Hold'}\\
      -\psi & \text{otherwise}
    \end{cases},
\end{equation}
where in $R(s^{o}_{t}, h^{o}_{t})$, $d^{c}_{o}$ is the distance from the ownship to the closest intruder aircraft and $d^{\text{MAX}_{i}}_{o}$ is the maximum distance to consider the closest intruder aircraft in the reward function. The hyperparameters $\alpha$ and $\delta$ are small, positive constants to penalize aircraft as they approach the separation threshold $d^{\text{NMAC}}$. In $R(a_{t})$, $\psi$ represents a penalty for advisories that require a deviation from the aircraft's current speed. This equates to minimizing speed changes, or minimize alerting of the algorithm. Finally, in $R(s^{o}_t,h^{o}_t,a_{t})$, the hyperparameter $\Omega$ represents a small, positive constant that is applied at every step in scenario. This parameter discourages aircraft from airborne holding, since slower aircraft will incur the $\Omega$ penalty for extended time. Table~\ref{use_case_param} displays the finalized hyperparameters used in this use-case.

\subsubsection{UAMToolkit Airspace Definition}
\begin{figure}[t]
\centering
\includegraphics[width=0.45\textwidth]{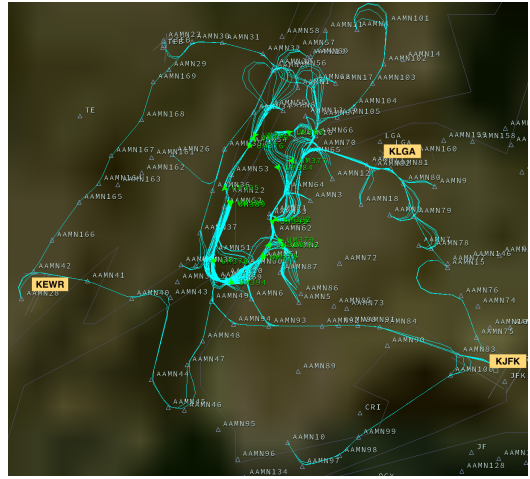}
\caption{Simulated AAM traffic in the NYC area displayed in the BlueSky air traffic simulator. Blue lines represent flown AAM trajectories. Green arrows represent AAM traffic currently in-flight.}
\label{bluesky_network}
\end{figure}

To compare the performance of algorithms a test scenario was generated using UAMToolkit. Near term AAM operations are expected to use existing helicopter routes as they mimic the concept of AAM corridors at lower densities~\cite{faa_conops}. As such, the scenario utilizes the VFR helicopter network presented in~\cite{doi:10.2514/6.2021-2381} with multiple lanes stacked vertically. The network was originally created as a replica of the VFR routes from the sectional charts of the New York airspace. During scenario generation, the aircraft requests a route to travel between the Origin Destination (OD) pair and is assigned an altitude within the route that is not already in use between the OD pair. This avoids head-on conflict, but it does not assure that another OD route or aircraft within the same OD route will be overtaken or free of conflict. The scenario generation takes into account imperfect altitude holding by the aircraft by adding noise in the form of a uniform distribution with a minimum value of -100 ft and maximum value of 100 ft to offset the selected altitude. The scenario is initiated in UAMToolkit with 100 aircraft total in the fleet that are trying to transport 1\% of the demand provided by the New York City Taxi Cab demand model. The simulation starts at midnight with all aircraft evenly distributed among the 29 vertiports in the network and continues for a full day of operations. Figure~\ref{bluesky_network} illustrates the flown AAM air traffic in the scenario for a day of simulation. Each vehicle is assigned a cruising speed property by a uniform distribution between 87 and 150 knots. Due to the ambiguity of takeoff and landing procedures for AAM aircraft, the BlueSky scenario spawns aircraft at the origin vertiport at the prescribed altitude and accelerates to its assigned cruising speed. Since the use-case will only impact the horizontal speed of the aircraft, the only vehicle properties of interest are the horizontal acceleration, minimum cruising speed, and maximum speed. For this use-case the aircraft model used in BlueSky was chosen as a Eurocopter EC-135 because its flight envelope encompassed the cruising assumptions of the UAMToolKit aircraft model. Thus, by using this EC-135 model, the simulation bounds the cruising speed of the aircraft regardless of whether the algorithm request the aircraft to speed up or slow down.

\begin{table}[tb]
\caption{AAM corridor separation assurance use-case hyperparameters.}
\centering
\label{use_case_param}
\begin{tabular}{lll}
Parameter                    & Value &  \\ \hline
\multicolumn{1}{l|}{NMAC}  & 150m   \\
\multicolumn{1}{l|}{LoWC} &  450m     &  \\
\multicolumn{1}{l|}{Desired speeds} &  [0, Maintain, 150] knots    &  \\
\multicolumn{1}{l|}{$N_{\text{wpt}}$} &  2    &  \\
\multicolumn{1}{l|}{$N_{\text{intruders}}$} &  30    &  \\
\multicolumn{1}{l|}{$d^{\text{MAX}}_{o}$} &  3000m     &  \\
\multicolumn{1}{l|}{Reward coefficient $\alpha$} &  1     &  \\
\multicolumn{1}{l|}{Reward coefficient $\delta$} &  0.0003     &  \\
\multicolumn{1}{l|}{Reward coefficient $\psi$} &  0.001     &  \\
\multicolumn{1}{l|}{Reward coefficient $\Omega$} &  0.001     &  \\
\hline
\end{tabular}
\end{table}

\subsection{Algorithms}

For this study, two algorithms were applied to the corridor separation assurance use-case; double deep Q-networks ~\cite{van_Hasselt_Guez_Silver_2016} and the deep distributed multi-agent variable with attention (D2MAV-A) algorithm~\cite{doi:10.2514/1.I010973}. In addition, a variant of D2MAV-A was also introduced and compared against the two algorithms.

\subsubsection{Deep Q-Networks}
Deep Q-Networks (DQN) is a general off-policy reinforcement learning algorithm that has been proven to achieve superior human-level performance in challenging games such as Atari~\cite{mnih2015human}. This generic algorithm can be applied to many different use-cases so long as the use-case can be formulated as a Markov Decision Process (MDP). In AAM-Gym, the Deep Q-Networks implementation is also parallel-enabled, meaning that parallel workers can be instantiated to improve the data collection frequency and training of the algorithm. Given that Deep Q-Networks is a generic algorithm with no input/output requirements, this algorithm can be applied to other use-cases developed in AAM-Gym.

Double Deep Q-Networks (DDQN) is an extension to DQN that was shown to reduce the overestimation bias in the Q-values that was present in DQN~\cite{van_Hasselt_Guez_Silver_2016}. The DDQN variant of DQN is also available to use in AAM-Gym and due to the improved performance of DDQN over DQN, DDQN was selected for the numerical experiments. In Table~\ref{ddqn_params}, the hyperparameters of DDQN and their associated value for the numerical experiments are provided.\footnote{More information on DDQN and associated hyperparameters can be found in \cite{van_Hasselt_Guez_Silver_2016}.}

\begin{table}[tb]
\caption{DDQN Hyperparameters.}
\centering
\label{ddqn_params}
\begin{tabular}{lll}
Parameter                    & Value &  \\ \hline
\multicolumn{1}{l|}{Batch size} & 512 &     \\
\multicolumn{1}{l|}{Nodes} &  128     &  \\
\multicolumn{1}{l|}{$\gamma$} &  0.99     &  \\
\multicolumn{1}{l|}{$\epsilon$ decay steps} &  500000     &  \\
\multicolumn{1}{l|}{$\epsilon_{\text{start}}$} &  0.999     &  \\
\multicolumn{1}{l|}{$\epsilon_{\text{end}}$} &  0.0001     &  \\
\multicolumn{1}{l|}{Replay memory size} &  5000000     &  \\
\multicolumn{1}{l|}{Learning rate} &  0.0001     &  \\
\multicolumn{1}{l|}{Target network update freq.} &  50000    &  \\ 
\multicolumn{1}{l|}{$N_{\text{workers}}$} &   40   &  \\ \hline
\end{tabular}
\end{table}

\subsubsection{D2MAV-A}

Deep distributed multi-agent variable with attention (D2MAV-A) is a state-of-the-art on-policy multi-agent reinforcement learning algorithm designed for autonomous self-separation in en route, traditional airspace~\cite{doi:10.2514/1.I010973}. D2MAV-A leverages attention networks~\cite{NIPS2017_3f5ee243} to handle variable state information from surrounding aircraft in the vicinity (i.e., intruder aircraft). Intruder aircraft are automatically assigned an importance weight by D2MAV-A, so that the algorithm can focus its ``attention'' on intruder aircraft that may violate separation requirements. This removes the need for defining a max number of intruder aircraft that D2MAV-A can process given that it is able to determine which intruder aircraft are most important and intelligently summarize the intruder aircraft's state into a fixed-length vector.

While D2MAV-A was designed for en route traditional airspace, there are no limitations for applying the algorithm to separation assurance in AAM corridors. In AAM-Gym, D2MAV-A represents a non-general algorithm, in comparison to DQN (or DDQN). D2MAV-A was designed specifically for separation assurance and would not be suitable for use-cases such as in-flight vertiport rerouting, for example. However, having support for use-case specific algorithms allows for unique AAM research studies where multiple AAM services can be deployed in the simulation. For example, a strategic deconfliction use-case can be developed with the option to have the aircraft receive tactical self-separation commands from D2MAV-A. This allows user the ability to understand how different AAM services interact once deployed. In Table~\ref{d2mav_params}, the hyperparameters of D2MAV-A and their associated value for the numerical experiments are provided.\footnote{More information on D2MAV-A and associated hyperparameters can be found in \cite{doi:10.2514/1.I010973}.}

\begin{table}[tb]
\caption{D2MAV-A Hyperparameters.}
\centering
\label{d2mav_params}
\begin{tabular}{lll}
Parameter                    & Value &  \\ \hline
\multicolumn{1}{l|}{Learning rate}  &   0.00001   \\
\multicolumn{1}{l|}{Batch size} &  512   &  \\
\multicolumn{1}{l|}{epochs} &  6   &  \\
\multicolumn{1}{l|}{Entropy coefficient $\beta$} &  0.0001   &  \\
\multicolumn{1}{l|}{PPO ratio bound $\epsilon$} &  0.1   &  \\
\multicolumn{1}{l|}{Nodes} &  128   &  \\
\multicolumn{1}{l|}{GAE discount factor $\gamma$} &  0.99   &  \\
\multicolumn{1}{l|}{GAE discount factor $\lambda$} &  0.95   &  \\
\multicolumn{1}{l|}{Leaky ReLU $\alpha$} &  0.2   &  \\
\multicolumn{1}{l|}{$N_{\text{workers}}$} &   40   &  \\ \hline
\end{tabular}
\end{table}

\subsection{Baselines}
The baseline used in this study was an unequipped aircraft that does not implement any self-separation commands. In this setting, the aircraft follows the assigned speeds and route given by the predeparture flight plan provided by UAMToolkit. This baseline enables users to understand how well algorithms perform relative to when there is no algorithm introduced.

\section{Results}
\label{results}
In this section, the performance of the DQDN, D2MAV-A with shared network layers (as introduced in \cite{doi:10.2514/1.I010973}), and D2MAV-A without shared network layers are compared on the separation assurance use-case with safety metrics provided by AAM-Gym. One scenario is developed for training which consists of 100 AAM aircraft operating over 25 minutes of operation time. This provides a challenging, high-density scenario for the algorithms. For evaluation, a second scenario is developed with 678 aircraft operating over 5 hours of operation time. This provides a more complex scenario, given this scenario was not used during training and the algorithms must generalize to unforeseen aircraft interactions. In addition, AAM-Gym wall-clock time training metrics are analyzed to understand the efficiency of the architecture in it's first release version. This allows future versions of AAM-Gym to baseline wall-clock training time against prior versions. All algorithms were allowed to train for 100 hours, with the best algorithm weights extracted for validation. 100 hours was selected as it serves as a reasonable length of time to obtain a good policy in D2MAV-A for performance comparisons. Given this paper is focused on the overall AAM-Gym architecture and analysis capabilities, rather than algorithm optimization, additional training time may lead to better performance.

\begin{figure*}[t]
\centering
\includegraphics[width=0.8\textwidth]{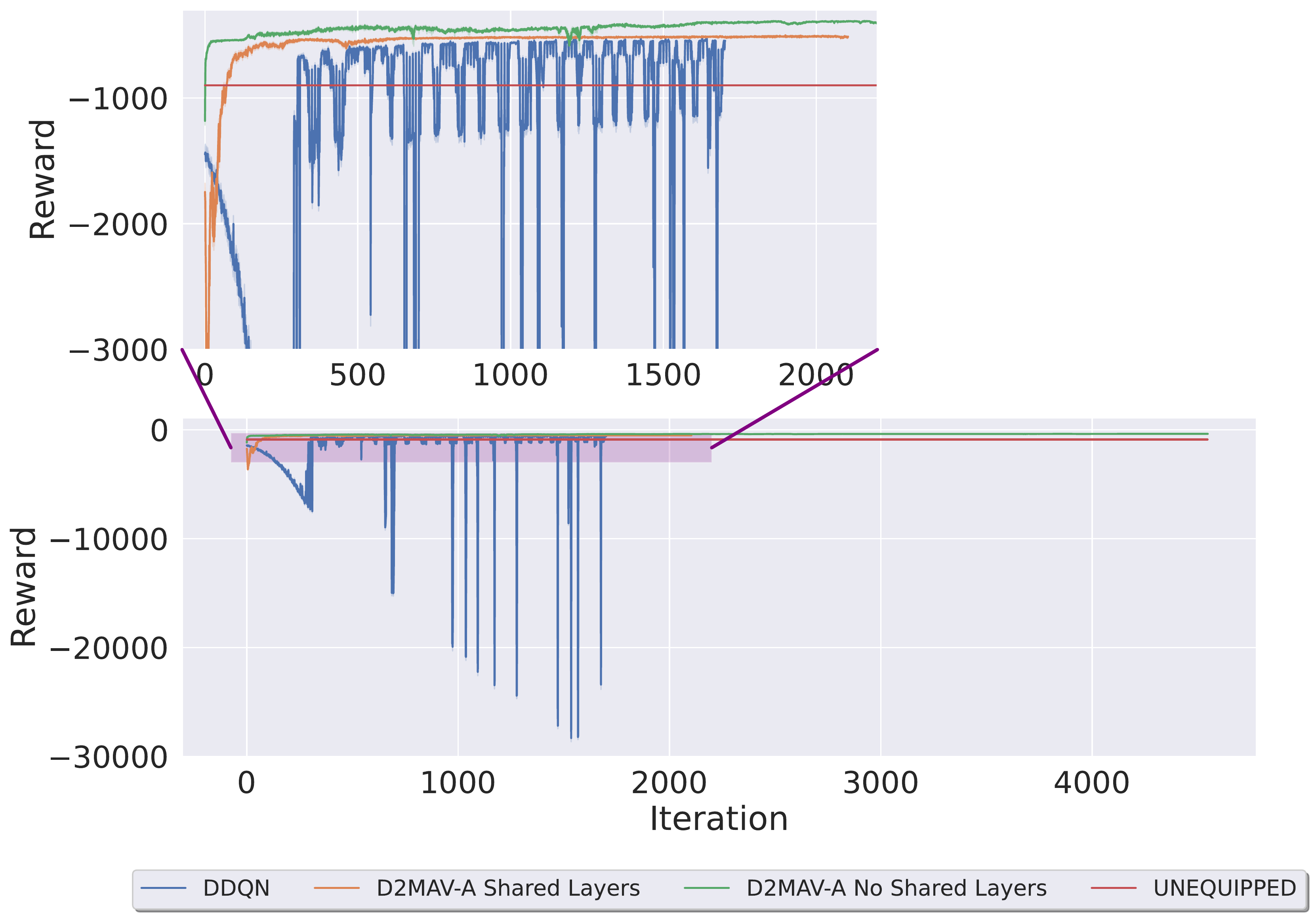}
\caption{Mean reward learning curve shown with 95\% confidence interval (lighter color). The full learning curve is displayed in the lower figure, with a zoomed in view of the shaded region displayed in the top figure for clarity. Each training iteration is composed of $N_{\text{workers}}$ simulation returns.}
\label{reward_curve}
\end{figure*}

\subsection{Learning Curve}
The convergence of the algorithms towards the optimal reward can be visualized through the learning curve. The learning curve represents the per-iteration returns received during the training phase (algorithms still updating internal logic). If an algorithm is able to be run in parallel (both DDQN and D2MAV-A can use parallel workers), then one training iteration represents the average return of $N_{\text{workers}}$ simulations. Figure~\ref{reward_curve} displays the learning curve for the separation assurance use-case. On the bottom figure in Figure~\ref{reward_curve}, the full learning curve is shown with the per-iteration reward, as defined in Equation~\ref{reward_func}, on the y-axis. Given that the DDQN algorithm uses a different exploration policy than D2MAV-A, there are periods where the performance of DDQN is significantly worse than both the unequipped aircraft and D2MAV-A algorithm. The top figure in Figure~\ref{reward_curve} represents a zoomed in view of the shaded region in the bottom figure to allow for a visual comparison in algorithm convergence. In this view, it can be seen that all algorithms are able to quickly outperform the unequipped aircraft, with both variants of the D2MAV-A algorithm consistently outperforming the unequipped aircraft. In general, D2MAV-A No Shared Layers is the best performing algorithm given it is achieving the maximum reward.

The range of values on the y-axis are indicative to the optimization objectives defined in Equation~\ref{reward_func}. Given the primary objective (separation assurance) is given the most weight, large changes in the reward are seen early on in training as the algorithms are focusing on optimizing this objective. After the algorithms learn to optimize the primary objective, the secondary objectives are then optimized, resulting in a slower, gradual increase in reward over training. This is due to the fact that the secondary objectives are given a smaller weight, or penalty since maintaining sufficient separation is the primary objective of the use-case. An important observation from Figure~\ref{reward_curve}, is that after 100 hours of training, the algorithms do not complete the same number of training iterations. One training iteration is completed at the conclusion of the 100 aircraft scenario, so depending on the algorithm's behavior, a scenario may take longer to complete. For example, if an algorithm is preferring airborne holding to maintain separation, the time to complete the scenario will take much longer since the training iteration will not complete until all aircraft have reached their destination vertiport. Therefore, it can be concluded that after 100 hours of training, the D2MAV-A No Shared Layers algorithm has learned a more efficient separation strategy that minimizes airborne holding.

\begin{figure}[t]
\centering
\includegraphics[width=\columnwidth]{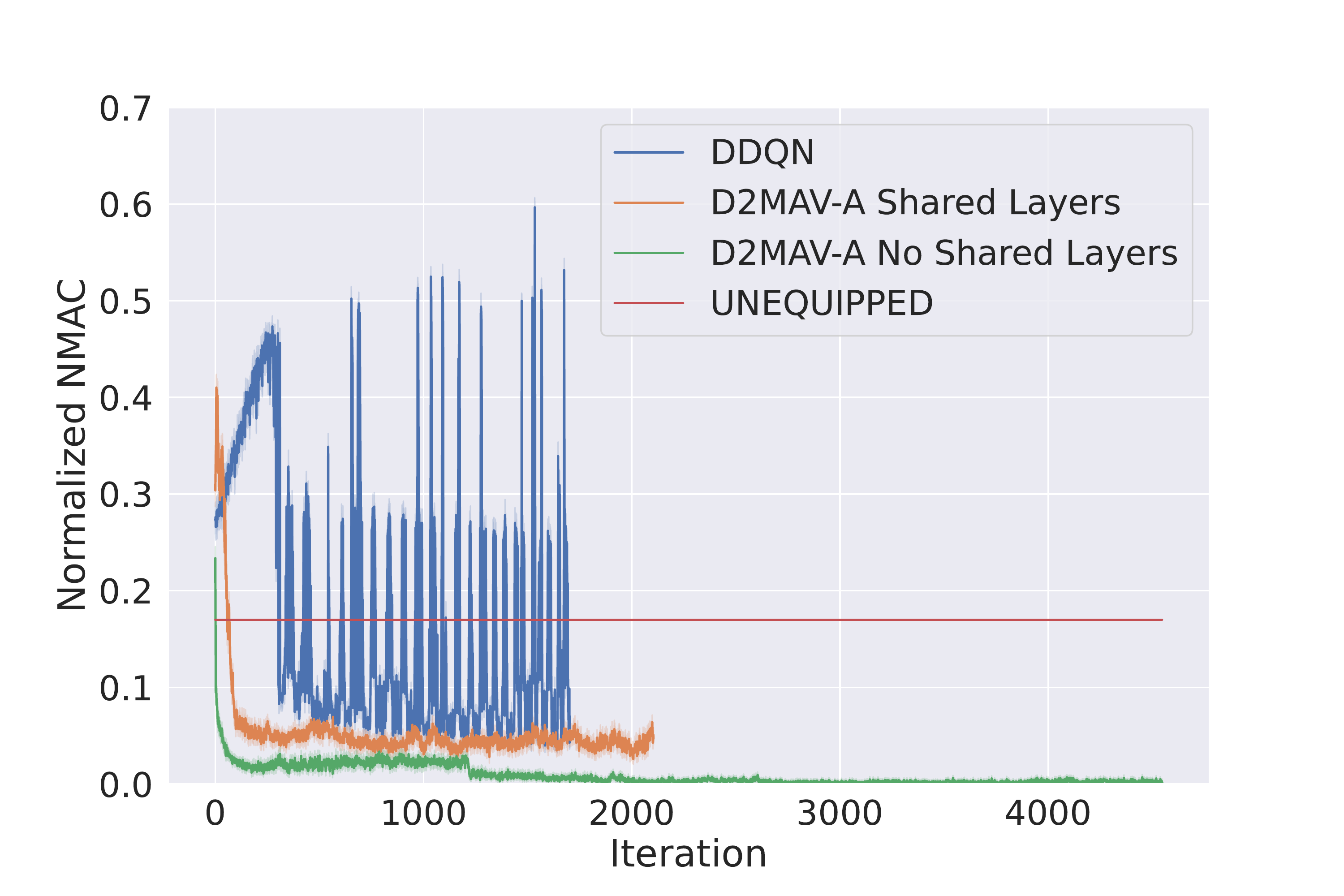}
\caption{Mean normalized NMAC learning curve shown with 95\% confidence interval. Each training iteration is composed of $N_{\text{workers}}$ simulation returns.}
\label{learning_curve}
\end{figure}

The learning curve can also show the algorithm convergence towards the primary objective in the reward function. Figure~\ref{learning_curve} displays the near mid-air collision count (NMAC), normalized by the number of aircraft in the scenario (100) over the training phase. In Figure~\ref{learning_curve}, best performance equates to minimizing the the normalized NMAC. It can be seen from Figure~\ref{learning_curve} that the minimization of NMAC is inversely correlated with the maximization of reward in Figure~\ref{reward_curve}, with the D2MAV-A No Shared Layers algorithm achieving the best performance. 

\subsection{Algorithm Validation}
After the training phase is complete, the best performing algorithm weights are extracted and evaluated with no additional algorithm logic updates for 100 evaluation iterations.\footnote{Best performance is based on a 25-iteration rolling mean of reward.} Validation is performed on two scenarios: (1) the 100 aircraft scenario used in training and (2) the 678 aircraft scenario not used in training.

The risk ratio is a standard metric used to evaluate tactical separation systems. Risk ratio is defined as the number of algorithm NMACs divided by the number of unequipped NMACs. This metric provides insight into the safety benefits a particular algorithm may have compared to the nominal case where all aircraft are unequipped without any separation logic. A risk ratio less than one represents increased safety, with a risk ratio greater than one representing decreased safety. A risk ratio equal to one represents safety no better than the that of the unequipped aircraft. Table~\ref{nmac_results} displays the risk ratio for each algorithm on both the 100 and 678 aircraft scenarios. It can be seen that on the 100 aircraft scenario, the D2MAV-A No Shared Layers algorithm significantly outperforms the D2MAV-A Shared Layers and DDQN algorithms, achieving a risk ratio of 0.0107. In the 678 aircraft scenario, all algorithms achieve safety improvement over the unequipped aircraft, however the risk ratios are much higher than the 100 aircraft scenario. In this case, the algorithms may need additional training time or a more complex training scenario.

The loss of well clear (LoWC) ratio is similar to the risk ratio, but is based on the number of LoWCs instead of NMACs. Table~\ref{lowc_results} displays the LoWC ratio evaluation results and similar conclusions to the risk ratio can be observed from the results. The D2MAV-A No Shared Layers outperforms the other algorithms on both the 100 and 678 aircraft scenarios. While a LoWC of 0.5 is large, an agreed upon standard for LoWC in AAM has yet to be defined. In AAM-Gym, the NMAC and LoWC standard is defined as a hyperparameter. In this way, AAM-Gym can be used to understand how AI algorithms may impact current and future separation standards.

\begin{table}[tbp]
\caption{Mean and standard deviation risk ratio results over 100 evaluation iterations.}
\label{nmac_results}
\resizebox{\columnwidth}{!}{%
\begin{tabular}{llll}
Algorithm                    & Risk Ratio - 100 & Risk Ratio - 678 &  \\ \hline
\multicolumn{1}{l|}{D2MAV-A No Shared Layers}  &   0.0107 $\pm$ 0.0343    &  0.4875 $\pm$ 0.0303 &  \\
\multicolumn{1}{l|}{D2MAV-A Shared Layers} &  0.2443 $\pm$ 0.146     & 0.5926 $\pm$ 0.0648 &  \\
\multicolumn{1}{l|}{DDQN} &   0.4706 $\pm$ 0.0    & 0.5694 $\pm$ 0.0 &  \\ \hline
\end{tabular}
}
\end{table}

\begin{table}[tbp]
\caption{Mean and standard deviation LoWC ratio results over 100 evaluation iterations.}
\label{lowc_results}
\resizebox{\columnwidth}{!}{%
\begin{tabular}{llll}
Algorithm                    & LoWC Ratio - 100 & LoWC Ratio - 678 &  \\ \hline
\multicolumn{1}{l|}{D2MAV-A No Shared Layers}  &   0.5009 $\pm$ 0.0083    &  0.7159 $\pm$ 0.0112 &  \\
\multicolumn{1}{l|}{D2MAV-A Shared Layers} &  0.6722 $\pm$ 0.0807     & 0.7808 $\pm$ 0.0266 &  \\
\multicolumn{1}{l|}{DDQN} &   0.6667 $\pm$ 0.0    & 0.8654 $\pm$ 0.0 &  \\ \hline
\end{tabular}
}
\end{table}

\subsection{AAM-Gym Wall-Clock Metrics}
Through algorithm training and the validation comparison on the separation assurance use-case, AAM-Gym has demonstrated its effectiveness for AI research in AAM. However, it is important to understand the computational complexity of running AAM-Gym and how much data can be collected for the algorithms. The results in this section are based upon the D2MAV-A No Shared Layers algorithm, given D2MAV-A processes all aircraft data from each training iteration, therefore providing a more representative wall-clock time metric for how much data AAM-Gym can provide.

Over the 100 training hours, 342,274,846 aircraft states were provided by AAM-Gym for algorithm optimization. Given that $N_{\text{workers}}=40$, each worker generated approximately 8,556,871 aircraft states over the 100 hours, or 24 aircraft states per second. The distributed computing capability of AAM-Gym provides an opportunity to massively scale $N_{\text{workers}}$. MIT Lincoln Laboratory's Supercomputing Center provides access to over 41,000 CPU cores for parallel processing~\cite{reuther2018interactive}. This enables up to 427,843,558 aircraft states to be generated per hour given 5,000 CPU cores (1 worker = 1 CPU core). In future work, a detailed trade-off analysis between algorithm performance and number of CPU cores will be performed to understand the performance impact of large scale distributed computing in AAM use-cases.

\section{Conclusion}
\label{conclusion}
In this paper, AAM-Gym is introduced as the first research and validation testbed for AI research in AAM. In addition to leveraging community standard open source modules and API's, AAM-Gym provides a suite of scenario generation tools to enhance the realism of AAM simulations. The full pipeline of AAM-Gym is validated on a separation assurance in AAM corridor use-case with two AI algorithms, DDQN and D2MAV-A. Through metrics provided by AAM-Gym, rapid analysis can be performed to reduce the time and complexity from algorithm development to airspace metrics analysis.

With the integration of real-world systems such as ACAS sXu, AAM-Gym provides a unique capability to baseline against, or evaluate AI algorithms with ACAS sXu. The feedback provided from this testbed enables detailed analyses of both airspace and aircraft level use-cases, along with applicability of AI in AAM. This can inform policy makers of key infrastructural gaps that need to be addressed in AAM, or uncover potential barriers to AI certification in air transportation. AAM-Gym has the potential to enable a broad community to actively engage in AAM research and foster collaboration through standardized use-cases and interfaces.



\end{document}